\documentclass[10pt,journal,compsoc]{IEEEtran}
\usepackage{url}
\usepackage{graphicx}
\usepackage{amsmath}
\usepackage{algorithm}
\usepackage{amsthm}
\usepackage{amsfonts}
\usepackage{makecell}
\usepackage{multirow}

\ifCLASSOPTIONcompsoc
  \usepackage[nocompress]{cite}
\else
  \usepackage{cite}
\fi
\ifCLASSINFOpdf
\else
\fi
\begin{document}
%
% paper title
% Titles are generally capitalized except for words such as a, an, and, as,
% at, but, by, for, in, nor, of, on, or, the, to and up, which are usually
% not capitalized unless they are the first or last word of the title.
% Linebreaks \\ can be used within to get better formatting as desired.
% Do not put math or special symbols in the title.
\title{Context-Sensitive Generation Network for Handing Unknown Slot Values in Dialogue State Tracking}
%
%
% author names and IEEE memberships
% note positions of commas and nonbreaking spaces ( ~ ) LaTeX will not break
% a structure at a ~ so this keeps an author's name from being broken across
% two lines.
% use \thanks{} to gain access to the first footnote area
% a separate \thanks must be used for each paragraph as LaTeX2e's \thanks
% was not built to handle multiple paragraphs
%
%
%\IEEEcompsocitemizethanks is a special \thanks that produces the bulleted
% lists the Computer Society journals use for "first footnote" author
% affiliations. Use \IEEEcompsocthanksitem which works much like \item
% for each affiliation group. When not in compsoc mode,
% \IEEEcompsocitemizethanks becomes like \thanks and
% \IEEEcompsocthanksitem becomes a line break with idention. This
% facilitates dual compilation, although admittedly the differences in the
% desired content of \author between the different types of papers makes a
% one-size-fits-all approach a daunting prospect. For instance, compsoc 
% journal papers have the author affiliations above the "Manuscript
% received ..."  text while in non-compsoc journals this is reversed. Sigh.

\author{Puhai~Yang,~\IEEEmembership{}
        ~Heyan~Huang,~\IEEEmembership{}
        and~Xian-Ling~Mao~\IEEEmembership{}% <-this % stops a space
\IEEEcompsocitemizethanks{\IEEEcompsocthanksitem P. Yang, H. Huang and X. Mao are with School
of  Computer Science and Technology, Beijing Institute of Technology, Beijing, China.\protect\\
% note need leading \protect in front of \\ to get a newline within \thanks as
% \\ is fragile and will error, could use \hfil\break instead.
E-mail: \{3120195507, hhy63, maoxl\}@bit.edu.cn}
%\IEEEcompsocthanksitem J. Doe and J. Doe are with Anonymous University.}% <-this % stops an unwanted space
%\thanks{Manuscript received xx xx, xxxx; revised xx xx, xxxx.}}
}
\IEEEtitleabstractindextext{%
\begin{abstract}
As a key component in a dialogue system, dialogue state tracking plays an important role. It is very important for dialogue state tracking to deal with the problem of unknown slot values. As far as we known, almost all existing approaches depend on pointer mechanism to solve the unknown slot value problem. These pointer mechanism-based methods usually have a hidden assumption that there is at most one out-of-vocabulary word in an unknown slot value because of the character of a pointer mechanism. However, often, there are multiple out-of-vocabulary words in an unknown slot value, and it makes the existing methods perform bad. To tackle the problem, in this paper, we propose a novel \textbf{C}ontext-\textbf{S}ensitive \textbf{G}eneration network (\textbf{CSG}) which can facilitate the representation of out-of-vocabulary words when generating the unknown slot value. Extensive experiments show that our proposed method performs better than the state-of-the-art baselines.
\end{abstract}

% Note that keywords are not normally used for peerreview papers.
}

% make the title area
\maketitle

% To allow for easy dual compilation without having to reenter the
% abstract/keywords data, the \IEEEtitleabstractindextext text will
% not be used in maketitle, but will appear (i.e., to be "transported")
% here as \IEEEdisplaynontitleabstractindextext when the compsoc 
% or transmag modes are not selected <OR> if conference mode is selected 
% - because all conference papers position the abstract like regular
% papers do.
\IEEEdisplaynontitleabstractindextext
% \IEEEdisplaynontitleabstractindextext has no effect when using
% compsoc or transmag under a non-conference mode.

% For peer review papers, you can put extra information on the cover
% page as needed:
% \ifCLASSOPTIONpeerreview
% \begin{center} \bfseries EDICS Category: 3-BBND \end{center}
% \fi
%
% For peerreview papers, this IEEEtran command inserts a page break and
% creates the second title. It will be ignored for other modes.
\IEEEpeerreviewmaketitle

\IEEEraisesectionheading{\section{Introduction}\label{sec:introduction}}
% Computer Society journal (but not conference!) papers do something unusual
% with the very first section heading (almost always called "Introduction").
% They place it ABOVE the main text! IEEEtran.cls does not automatically do
% this for you, but you can achieve this effect with the provided
% \IEEEraisesectionheading{} command. Note the need to keep any \label that
% is to refer to the section immediately after \section in the above as
% \IEEEraisesectionheading puts \section within a raised box.

% The very first letter is a 2 line initial drop letter followed
% by the rest of the first word in caps (small caps for compsoc).
% 
% form to use if the first word consists of a single letter:
% \IEEEPARstart{A}{demo} file is ....
% 
% form to use if you need the single drop letter followed by
% normal text (unknown if ever used by the IEEE):
% \IEEEPARstart{A}{}demo file is ....
% 
% Some journals put the first two words in caps:
% \IEEEPARstart{T}{his demo} file is ....
% 
% Here we have the typical use of a "T" for an initial drop letter
% and "HIS" in caps to complete the first word.
\IEEEPARstart{C}{urrently}, the research and application of dialogue systems are widely concerned, especially for task-oriented dialogue systems, such as booking tickets and ordering restaurants. Dialog state tracking is a key component of a task-oriented dialogue system. By parsing dialogue history, dialog state tracking extracts user’s intentional state, such as intention, slot and value, as the input of dialogue manager for system decision making. For example, (price, cheap) and (area, centre) are extracted from “I am looking for a cheap restaurant in the centre of the city” as user’s state.

\begin{figure}[h]
	\centering
	\includegraphics[width=\linewidth]{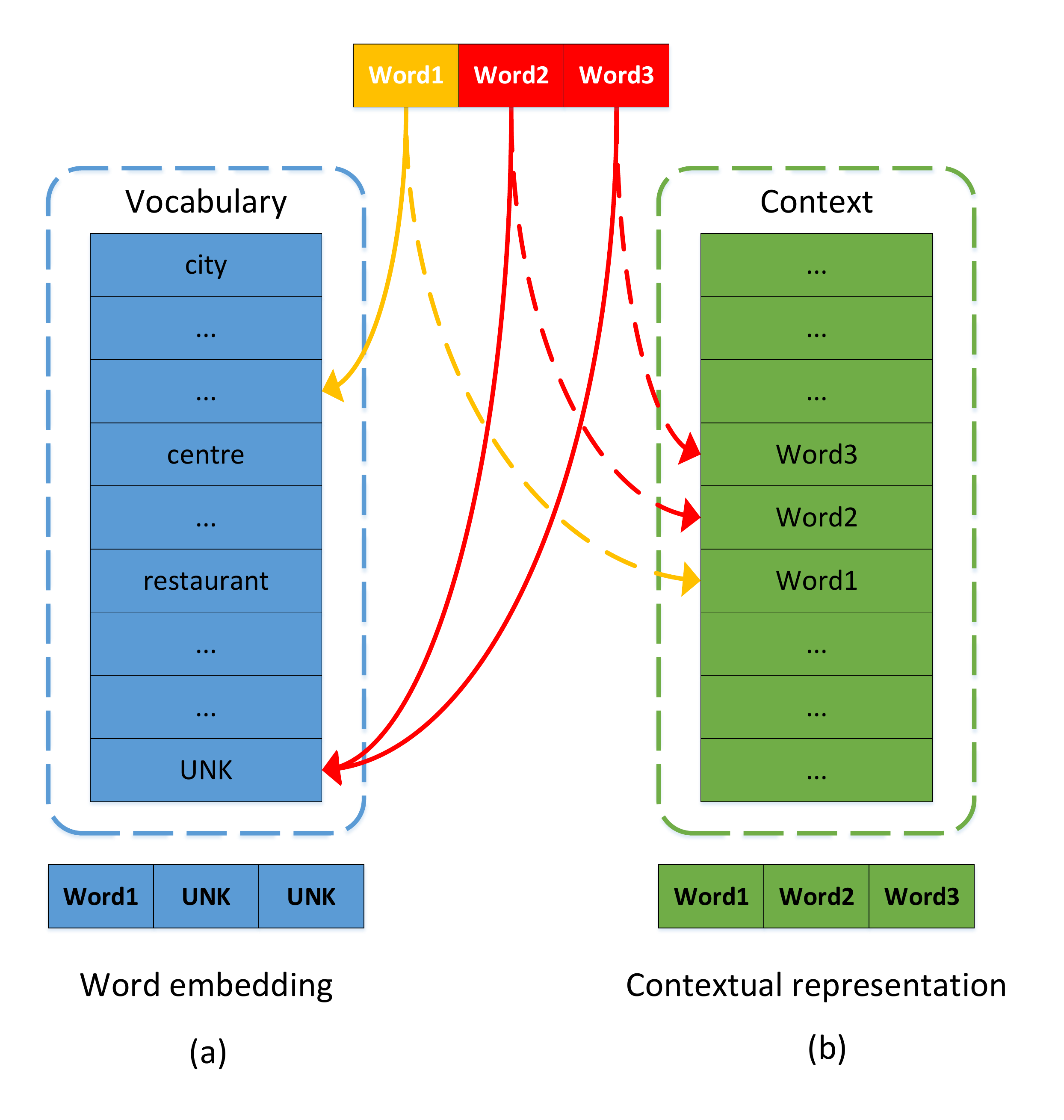}
	\caption{Two different representations of words: word embedding and contextual representation. The yellow unit is the word in the vocabulary, the red unit is out-of-vocabulary word, the blue unit refers to the vocabulary space representation of the word, namely word embedding, and the green unit refers to the representation of the word in a specific context.}
	%\Description{The 1907 Franklin Model D roadster.}
	\label{img1}
\end{figure}

Traditionally, dialog state tracking is typically solved using ontology-based approach \cite{yazdani2015model}\cite{mrkvsic2017neural}\cite{zhong2018global},  which is based on the assumption that all slot values are known in advance. In reality, however, it is impossible to know all the slot values, dialog state tracking models often encounter slot values that have never been seen in training, which are also known as unknown slot values \cite{xu2018end}. Thus, recently, researches on dialog state tracking mainly concentrates on the open-vocabulary dialog state tracking method \cite{zhao2018improving}\cite{zhao2019hierarchical}\cite{wu2019transferable}\cite{huang2019streamlined}, which attempt to solve the problem of unknown slot values by generating novel words through vocabulary-based distribution.

However, the open-vocabulary dialog state tracking method almost always depends on pointer mechanism \cite{vinyals2015pointer} to extract unknown slot values on account of the fact that unknown slot values contain the out-of-vocabulary word. And the validity of pointer mechanism to extract unknown slot values is usually based on the assumption that the unknown slot value contains not more than one out-of-vocabulary word. Invariably, in the decoder, pointer mechanism use word embedding to represent each word, and all out-of-vocabulary words are represented by a uniform embedding, such as ”UNK”. So that multiple out-of-vocabulary words in an unknown slot value are indistinguishable, as shown in Fig \ref{img1} (a), which confuses the decoder that uses the word embedding as input.

In fact, there is often more than one out-of-vocabulary words in an unknown slot value, the pointer mechanism cannot distinguish these different out-of-vocabulary words only by word embedding, and the information of these out-of-vocabulary words cannot be adequately represented by a uniform embedding. Due to the input uncertainty that comes with this situation during decoding, the output of the decoder will gradually deviate, resulting in the error of the unknown slot value.

To tackle the drawback, we emphasize that the input of the decoder should be infused with more information than just the word embedding. In this paper, we propose a novel Context-Sensitive Generation network (CSG) for the unknown slot value problem in dialog state tracking. Our proposed model joins word contextual information, as shown in Fig \ref{img1} (b), to the input of the decoder. So that different out-of-vocabulary words can be distinguished by word contextual information, and word contextual information can also enrich the representation of the word, which used to be represented only by word embedding.

The main contributions of this paper are as follows:
\begin{itemize}
	\item We propose a novel context-sensitive generation network that utilizes word contextual information to overcome the problem of uncertain information caused by out-of-vocabulary words.
	\item Our model is highly portable and can be conveniently deployed to existing pointer-mechanism-based dialog state tracking models.
	\item On the most influential MultiWOZ 2.1 and DSTC2 benchmarks, our model has obvious advantages over the state-of-the-art baselines in the extraction of unknown slot values while maintaining the ability to know slot values.
\end{itemize}

The rest of paper is organized as follows: Related work is briefly introduced in section 2 and the shortcomings are pointed out. In section 3, The main structure of our proposed model is described in detail. Experiments and analysis are presented in section 4, followed by conclusions in section 5.

% You must have at least 2 lines in the paragraph with the drop letter
% (should never be an issue)

%I wish you the best of success.

%\hfill mds
 
%\hfill August 26, 2015

\section{Related Work}

As we mentioned above, the open-vocabulary dialog state tracking method usually depends on pointer mechanism \cite{vinyals2015pointer} to solve the unknown slot value problem when in dialog state tracking (DST). The validity of pointer mechanism implies the assumption that slot value contains not more than one out-of-vocabulary word, which is the motivation of this paper. 

Different from the traditional ontology-based DST method \cite{williams2012a}\cite{ren2018towards}\cite{ramadan2018large}\cite{lee2019sumbt}, which faces the problem of incomplete ontology due to limited knowledge and resources in practical application, open-vocabulary DST \cite{xu2018end}\cite{gao2019dialog}\cite{Kim2019EfficientDS}\cite{ren2019scalable}\cite{Zhu2020EfficientCA} depends on Encoder-Decoder structure and attention mechanism to generate dialog state directly from dialog history. For the problem of unknown slot value, the most commonly adopted strategies are pointer network \cite{vinyals2015pointer} and pointer-generator networks \cite{see2017get}. The former directly extracts the value of each slot from the dialogue history, while the latter generates or extracts slot value by combining the dialogue history distribution and vocabulary-based distribution. Next, we introduce some dialogue state tracking models that employ these two strategies for dealing with unknown slot values.

\subsection{Extractive DST}

The Encoder-Decoder structure based on pointer network has been used in many researches \cite{wang2016machine}\cite{jadhav2018extractive}\cite{li2019adversarial}. These models are designed to directly copy the words in the input text, which is different from the traditional generative method of modeling vocabulary distribution, but more like a variant of sequence labeling \cite{kim2014sequential}. Since almost all slot values are contained in the dialogue history, pointer network-based extractive DST was proposed for DST modeling \cite{xu2018end}, and the extractive DST can also solve the unknown slot value problem. This method can copy slot values directly from the dialogue history, but it faced with the problem of subsequent processing, because some slot values need to be inferred instead of being directly contained in the dialogue history.

SpanPtr \cite{xu2018end} first introduce the pointer network into dialogue state tracking to solve the problem of unknown slot values. The model in SpanPtr consists of a bidirectional RNN to encode dialogue history and a unidirectional RNN to decode slot values, and a gate mechanism is applied to determine whether slot values need to be decoded. During the decoding, SpanPtr relies on the pointer network to generate the  dialogue history distribution to obtain the starting and ending positions of slot values in the dialogue history, in this way, the unknown slot values that do not exist in the word vocabulary can also be extracted from the dialogue history.

\subsection{Hybrid DST}

\begin{figure*}[h]
	\centering
	\includegraphics[width=\linewidth]{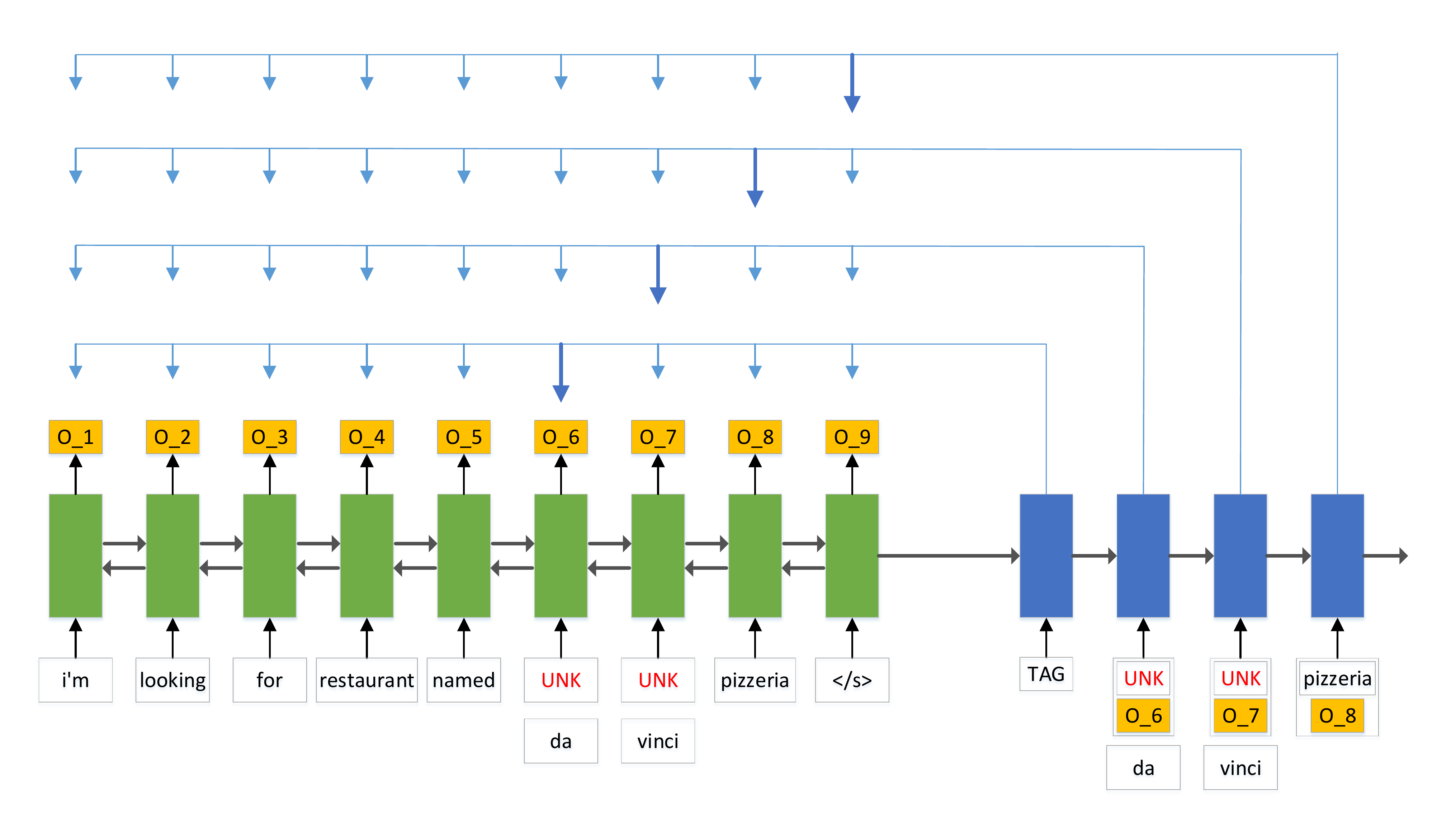}
	\caption{The framework of our proposed model. The green unit is the encoder, the blue unit is the decoder, the
		yellow unit is the output vector of the encoder for each encoding step, ”da” and ”Vinci” are out-of-vocabulary
		words, and the word ”UNK” is used as the substitution of out-of-vocabulary words. TAG means a slot, such as
		”restaurant-name”.}
	%\Description{The 1907 Franklin Model D roadster.}
	\label{img2}
\end{figure*}

Pointer-Generator Networks (PGN) \cite{see2017get}, a hybrid between sequence-to-sequence attentional model \cite{nallapati2016abstractive} and a pointer network, has received a lot of attention since it was proposed, and there are also many relevant studies in DST \cite{zhao2018improving}\cite{wu2019transferable}\cite{huang2019streamlined}. Different from the extractive method, PGN-based DST, hybrid DST we called, can copy words from the input text via pointer network while maintaining the ability to generate novel words using vocabulary-based distribution. Therefore, the hybrid DST does not require subsequent processing modules.

HD \cite{zhao2019hierarchical} is a hierarchical model for spoken language understanding tasks, which takes into account the hierarchical relationship between action, slot and value. This hierarchical modeling can also be directly introduced into DST, which is modified as a strong baseline in this paper. HD consists of four parts, including a dialogue text encoder, an action classifier, a slot classifier, and a slot value generator based on PGN. In the inference phase, the model transmits information according to the action-slot-value hierarchy, and finally obtains the results of the three in turn.

TRADE \cite{wu2019transferable} is an improved variant of SpanPtr that uses PGN instead of the original pointer network. TRADE relies on PGN to generate each word in turn in its slot value decoder, thus freeing itself from the limitation of extracting slot values only from the dialogue history. In addition, TRADE treats the domain-slot tuple as a new slot, which enables the model to share slot parameters across domains, greatly improving the performance of the model on cross-domain data sets.

In order to reduce the scale of the dialog state that needs to be generated in each round of multi-round dialogues, SOM-DST \cite{Kim2019EfficientDS} propose to treat the dialog state as a fixed-sized memory, and then maintain the memory content in each round of dialogues. They divide the memory state into four categories: CARRYOVER, DELETE, DONTCARE, and UPDATE, which respectively represent the memory operations of copying the previous dialog state, changing the dialog state to none, changing the dialog state to dontcare, and generating a new dialog state. The sequence decoding method based on PGN is adopted to generate the new dialog state.

In general, current open-vocabulary DST methods are basically depend on pointer mechanism to solve unknown slot value problem. However, as we mentioned in the previous section, pointer mechanism is faced with the problem of uncertain input information in decoding. When there are multiple out-of-vocabulary words in an unknown slot value, the unknown slot value generated by the pointer in pointer mechanism will be deviated.

\section{Context-Sensitive Generation Network}

In this section, we describe (1) the framework of our proposed model and (2) different schemes to leverage context information in our model. The code for this paper is available online\footnote{\url{https://github.com/yangpuhai/CSG}}.

\subsection{Framework}

In our model (shown in Fig \ref{img2}) the encoder is used to generate the vector representation of dialogue history and the contextual representation of each word. It is important to note that the encoder can be any encoding model, such as bi-LSTM \cite{hochreiter1997long} and bi-GRU \cite{cho-etal-2014-learning}. The input of the encoder is the dialogue history $H=[w_1,...,w_l]\in{\mathbb{R}^{l\times{d_{emb}}}}$, which is the concatenation of all words in the dialogue history. $l$ is the length of dialogue history and $d_{emb}$ is the size of word embedding. The output of the encoder consists of two parts, one is the hidden
state $S\in{\mathbb{R}^{d_{hid}}}$ at the last encoding step, which is the initial input state of the decoder, and the other is the output $OT=[o_1^{enc},...,o_l^{enc}]\in{\mathbb{R}^{l\times{d_{hid}}}}$ consisting of the output of the encoder at each encoding step, $d_{hid}$ is the hidden size. In our model, we assume that $OT$ is not only a representation of the dialogue history, but also a contextual representation of each word in the dialogue history. Therefore, $OT$ can be used to enhance the representation of out-of-vocabulary word in decoding.

At the initial step of decoding, we use the slot embedding as input to the decoder. It is important to point out that the slot embedding does not have to be the input to the decoder, but can also be placed in the output, which is not the focus of this paper. At decoding step $t$, the output $o_t^{dec}\in{\mathbb{R}^{d_{hid}}}$ of the decoder is used to generate the attention $p_t^{history}\in{\mathbb{R}^l}$ over each word in dialogue history.

$$p_t^{history}=softmax(OT\cdot{(o_t^{dec})^T})$$ 

Then, the position $pos_t$ of the $t_{th}$ word of slot value in the input dialogue history is determined by the maximum attention in $p_t^{history}$.

$$pos_t=\mathop{\arg\max}_{i\in{[1,...,l]}}p_t^{history}$$

At decoding step $t+1$, The pointer network traditionally take the embedding $w_{pos_t}$ of the word selected in step $t$ as the input $I_{t+1}$ of the decoder.
However, the information in the word embedding of the out-of-vocabulary word is incomplete and cannot effectively represent the word. As we mentioned above, the output $OT$ of the encoder can be seen as the contextual representation of each word in dialogue history. Therefore, our proposed model combines the embedding of word and its contextual representation as input to the decoder.

$$I_{t+1}=com(w_{pos_t},o_{pos_t}^{enc})$$

Where $com$ is the way $w_{pos_t}$ and $o_{pos_t}^{enc}$ are combined, and we will discuss the different combination schemes in the following.

\subsection{Context Utilization Schemes}

In this paper, we believe that words should have not only vocabulary space representation, that is, word embedding $w_{pos_t}$, but also contextual representation
$o_{pos_t}^{enc}$. In the traditional Encoder-Decoder model, only word embedding is usually considered and contextual representation is ignored. When a word is an out-of-vocabulary word, it is common to use a uniform word embedding “UNK” to represent the word. In this way, the information of the word cannot be adequately represented, which will lead to the deviation of the results. Therefore, we propose to combine the embedding and contextual representation of words, so that not only the information in words is enhanced, but also the unknown slot value problem can be effectively addressed.

In order to make effective use of word contextual information, we propose different schemes combining word contextual information with word embedding, as follows:

\textbf{Enc:} The contextual representation is used directly as the representation of the word. We directly change the input of the decoder to the context representation of the word. Without the learning of word embedding in the decoder, the model pays more attention to the modeling of the  dialogue history and the representation of each word in the specific context is more consistent, which may promote the solution of the dialogue history distribution in the extractive DST.

$$I_{t+1}=o_{pos_t}^{enc}$$

\textbf{Sum:} The sum of the word embedding and contextual representations is used as the representation of the word. We believe that in the existing hybrid DST, the input and output of the decoder do not match when it decodes the value of each slot, because the input is only vocabulary-based word embedding but the output contains two distributions: vocabulary-based distribution and dialogue history distribution. Therefore, we add the context representation of the word into the input of the decoder to enhance the information of the decoder when generating the dialogue history distribution. Meanwhile, the representations of different out-of-vocabulary words are also unique.

$$I_{t+1}=w_{pos_t}+o_{pos_t}^{enc}$$

\textbf{Cat:} The concatenation of the word embedding and contextual representations is used as the representation of the word. In the "Sum" scheme mentioned above, we directly sum the word embedding and the word contextual representation, which is an artificial way of combining information. However, the decoder in the hybrid DST may need the exact information from the corresponding input when generating the two distributions, rather than the information that has been artificially fused. Therefore, we concatenate the word embedding and the word contextual representation directly as the input of the decoder, so that the model can learn how to use the two kinds of information autonomously.

$$I_{t+1}=[w_{pos_t},o_{pos_t}^{enc}]$$

The above three word context utilization schemes are to explore different ways of context utilization in different situations. In a specific dialogue state tracking system, the implementation of the model is basically heterogeneous, so different context utilization schemes may have different effects. Next, we  elaborate on it through experimental analysis.

\begin{table}[!t]
	%% increase table row spacing, adjust to taste
	%\renewcommand{\arraystretch}{1.3}
	% if using array.sty, it might be a good idea to tweak the value of
	% \extrarowheight as needed to properly center the text within the cells
	\caption{Data statistics of DSTC2, MultiWOZ 2.1 and our Modified MultiWOZ 2.1.}
	\label{table1}
	\centering
	%% Some packages, such as MDW tools, offer better commands for making tables
	%% than the plain LaTeX2e tabular which is used here.
	\begin{tabular}{|c|c|c|c|}
		\hline
		Metric & DSTC2 & MultiWOZ 2.1 & \makecell[c]{Modified \\ MultiWOZ 2.1} \\
		\hline
		Domains & 1 & 5 & 5 \\
		\hline
		Slots & 3 & 30 & 7 \\
		\hline
		Values & 83 & 988  & 492  \\
		\hline
		Avg lengths per value & 1.04 & 1.82  & 2.58  \\
		\hline
		Avg states per turn & 2.15 & 5.41  & 1.88  \\
		\hline
		Training turns & 11,677 & 56,668  & 37,983  \\
		\hline
		Development turns & 3,934 & 7,374  & 5,431  \\
		\hline
		Test turns & 9,890 & 7,368  & 5,568  \\
		\hline
	\end{tabular}
\end{table}

\begin{table*}[!t]
	%% increase table row spacing, adjust to taste
	%\renewcommand{\arraystretch}{1.3}
	% if using array.sty, it might be a good idea to tweak the value of
	% \extrarowheight as needed to properly center the text within the cells
	\caption{The statistics of the modified MultiWOZ 2.1 in different out-of-vocabulary ratios. USV-O refers
		to the unknown slot value containing only one out-of-vocabulary word, while USV-M refers to the unknown slot
		value containing two or more out-of-vocabulary words.}
	\label{table2}
	\centering
	%% Some packages, such as MDW tools, offer better commands for making tables
	%% than the plain LaTeX2e tabular which is used here.
	\begin{tabular}{|c|c|c|c|c|c|c|c|c|c|c|c|}
		\hline
		\multirow{2}*{} & \multicolumn{11}{c|}{out-of-vocabulary ratios (\%)} \\
		\cline{2-12} & 0 & 10 & 20 & 30 & 40 & 50 & 60 & 70 & 80 & 90 & 100\\
		\hline
		USV-O in test set (\%) & 0 & 10 & 21 & 29 & 32 & 30 & 50 & 49 & 47 & 45 & 43\\
		\hline
		USV-M in test set (\%) & 0 & 2  & 3  & 12 & 18 & 26 & 35 & 38 & 45 & 51 & 57\\
		\hline
	\end{tabular}
\end{table*}

\section{Experiments}

\subsection{Dataset}

Our experiments are conduct on DSTC2 \cite{henderson2014second} and MultiWOZ 2.1 \cite{eric2019multiwoz}. DSTC2 is the traditional standard dialog state tracking benchmark on a single restaurant domain. In DSTC2, the dialog system interact with a user who want to find a specified restaurant around Cambridge,UK. The user can constrain the restaurant search by three informable slot: food type, area and price. We use system transcription, user transcription text, and goal-labels in DSTC2, and the training/development/test dataset contains 1612/506/1117 dialogs respectively.

MultiWOZ 2.1 dataset is the latest corrected version of the MultiWOZ dataset \cite{budzianowski2018multiwoz}. Compared with the DSTC2 dataset, MultiWOZ 2.1 containing around 10K dialogues, with each dialogue averaging 6.85 turns. And there are more than 30 slots and over 988 possible slot values in MultiWOZ 2.1. More importantly, since there are slot values containing multiple words in MultiWOZ 2.1 dataset, that is consistent with the problem of extracting unknown slot value containing multiple out-of-vocabulary words studied in this paper, so MultiWOZ 2.1 dataset is selected as benchmark. Further, to evaluate the ability of the model to extract unknown slot values, we create a modified MultiWOZ 2.1 dataset, we eliminate the slots in MultiWOZ 2.1 whose slot value contains only one word, and the final modified dataset contain 7 slots: ’traindestination’, ’train-departure’, ’attraction-name’, ’restaurant-name’, ’hotel-name’, ’taxi-destination’, ’taxi-departure’ in 5 domains: ”train”, ”attraction”, ”restaurant”, ”hotel”, ”taxi”. We list the details of DSTC2, MultiWOZ 2.1 and modified MultiWOZ 2.1 as shown in table \ref{table1}.

It should be noted that the slot values of the development and test sets of the modified MultiWOZ 2.1 dataset do not contain the unknown slot value. For experimental investigation, we select some words from the slot values of the development and test sets as out-of-vocabulary words to simulate the unknown
slot value problem. Specifically, we randomly select the word in the slot values from the development set and test set in different proportions, and then discard the word from the training vocabulary. Meanwhile, any sample containing the word in the training set changes the word to the character ”UNK”, but keeps the sample for training purposes. In order to highlight the experimental comparison, we discard the negative samples that do not contain any slot values in the data set without changing the experimental conclusion. The statistics of the modified MultiWOZ 2.1 in different out-of-vocabulary ratios are shown in Table \ref{table2}. Importantly, the out-of-vocabulary ratio mentioned in this paper refers to the ratio of out-of-vocabulary words in all slot values in the development and test sets.

\begin{table*}[!t]
	%% increase table row spacing, adjust to taste
	%\renewcommand{\arraystretch}{1.3}
	% if using array.sty, it might be a good idea to tweak the value of
	% \extrarowheight as needed to properly center the text within the cells
	\caption{Joint accuracy of dialog state tracking on MultiWOZ 2.1, DSTC2 and modified MultiWOZ 2.1. As an illustration of the name of our model, for example, SpanPtr\_CSG(Enc) refers to the improved DST model after adding the context-sensitive generation network we proposed into SpanPtr, where the context utilization scheme is ”Enc”.}
	\label{table3}
	\centering
	%% Some packages, such as MDW tools, offer better commands for making tables
	%% than the plain LaTeX2e tabular which is used here.
	\begin{tabular}{|c|c|c|c|c|c|}
		\hline
		\multirow{2}*{} & \multicolumn{2}{c|}{MultiWOZ 2.1} & \multirow{2}*{DSTC2} & \multicolumn{2}{c|}{Modified MultiWOZ 2.1}\\
		\cline{2-3}\cline{5-6} & 5 domains & Only restaurant &  & 5 domains & Only restaurant \\
		\hline
		SpanPtr           & 29.32 & 49.72 & 72.94 & 63.20 & \textbf{89.25} \\
		SpanPtr\_CSG(Enc) & \textbf{29.68} & 49.45 & \textbf{74.45} & \textbf{63.38} & \textbf{89.25} \\
		SpanPtr\_CSG(Sum) & 29.63 & \textbf{51.68} & 71.10  & 62.59 & 89.18 \\
		SpanPt\_CSG(Cat)  & 29.15 & 49.66 & 72.94  & 63.16 & 89.18 \\
		\hline
		SeqPtr           & 30.46 & 50.31 & 73.96  & \textbf{64.83} & \textbf{90.73} \\
		SeqPtr\_CSG(Enc) & \textbf{30.54} & 50.61 & 71.71  & 63.52 & 90.17 \\
		SeqPtr\_CSG(Sum) & 29.71 & \textbf{51.44} & 72.33  & 64.76 & 90.48 \\
		SeqPtr\_CSG(Cat) & 30.29 & 50.94 & \textbf{75.03}  & 64.03 & 90.48 \\
		\hline
		HD           & \textbf{36.14} & 62.79 & 75.59  & \textbf{66.25} & 90.42 \\
		HD\_CSG(Enc) & 31.19 & 60.70 & 75.47  & 65.85 & 92.40 \\
		HD\_CSG(Sum) & 36.09 & 64.42 & 73.35  & 65.64 & \textbf{92.95} \\
		HD\_CSG(Cat) & 33.85 & \textbf{64.51} & \textbf{79.22}  & 65.84 & 91.59 \\
		\hline
		TRADE           & 42.81 & \textbf{61.89} & 76.73  & 65.84 & 92.52 \\
		TRADE\_CSG(Enc) & 42.32 & 61.42 & \textbf{76.78}  & 65.73 & 91.66 \\
		TRADE\_CSG(Sum) & 43.72 & 61.63 & 75.63  & \textbf{67.37} & \textbf{92.66} \\
		TRADE\_CSG(Cat) & \textbf{43.95} & 59.12 & 73.67  & 65.28 & 92.65 \\
		\hline
	\end{tabular}
\end{table*}

\subsection{Baselines}

It is mentioned in section 2 that existing open-vocabulary dialog state tracking (DST) models are mainly divided into two types, pointer network-based extractive
DST and pointer-generator networks (PGN)-based hybrid DST. Therefore, our baselines include two types, the extractive model: SpanPtr \cite{xu2018end} and SeqPtr, 
and the hybrid model: HD \cite{zhao2019hierarchical} and TRADE \cite{wu2019transferable}. Next, we give a brief introduction to these models:

\textbf{SpanPtr:} This model uses pointer network to generate the start and end positions of slot values in a dialogue, and then extracts the slot values by copying.

\textbf{SeqPtr:} This is our modified version of the SpanPtr, this model generates the position of each word in the slot value in the dialogue instead of just the start and end positions.

\textbf{HD:} Hierarchical structure is considered in this model, where multiple classifiers are used to predict the existence of each slot, and then the slot
information is used to generate the slot value with PGN.

\textbf{TRADE:} This is the current state-of-the-art model on the MultiWOZ dataset. It uses a slot gate to predict whether slot values need to be generated, and there is a PGN-based state generator in the model to generate slot values.

All baselines and our models are set with the same parameters. Bi-GRU and GRU are used as encoder and decoder respectively. The dimension of word embedding and hidden state are both 400, and the dropout ratio is set to 0.2. All models are trained using the Adam optimizer with a batch size of 32, and all training consists of 50 epochs with early stopping on the validation set. In addition, word dropout is used on all models to improve generalization. More importantly, teacher forcing \cite{williams1989learning} with ratio of 0.5 is adopted by all models in decoding, except for the word contextual representation on PGN-based DST, in order to be consistent with baseline.

\subsection{Results}

\begin{table*}[!t]
	%% increase table row spacing, adjust to taste
	%\renewcommand{\arraystretch}{1.3}
	% if using array.sty, it might be a good idea to tweak the value of
	% \extrarowheight as needed to properly center the text within the cells
	\caption{Joint accuracy of dialog state tracking in multiple domains with different out-of-vocabulary ratios on
		modified MultiWOZ 2.1 dataset.}
	\label{table4}
	\centering
	%% Some packages, such as MDW tools, offer better commands for making tables
	%% than the plain LaTeX2e tabular which is used here.
	\begin{tabular}{|c|c|c|c|c|c|c|c|c|c|c|c|}
		\hline
		\multirow{2}*{} & \multicolumn{11}{c|}{out-of-vocabulary ratios (\%)} \\
		\cline{2-12} & 0 & 10 & 20 & 30 & 40 & 50 & 60 & 70 & 80 & 90 & 100\\
		\hline
		SpanPtr           & 63.2 & 62.8 & 61.8 & 60.8 & 59.5 & \textbf{60.0} & \textbf{59.7} & 56.7 & 54.8 & 54.1 & 47.5\\
		SpanPtr\_CSG(Enc) & \textbf{63.4} & \textbf{64.1} & 62.8 & \textbf{62.4} & \textbf{61.2} & 59.4 & 58.3 & \textbf{57.9} & 55.8 & 54.5 & 49.9\\
		SpanPtr\_CSG(Sum) & 62.6 & 61.9 & 62.9 & 60.3 & 61.0 & 59.1 & 58.2 & \textbf{57.9} & 55.8 & 53.8 & 51.2\\
		SpanPt\_CSG(Cat)  & 63.2 & 62.4 & \textbf{63.2} & 62.1 & 60.5 & \textbf{60.0} & 59.2 & 57.2 & \textbf{56.1} & \textbf{55.5} & \textbf{52.4}\\
		\hline
		SeqPtr           & \textbf{64.8} & \textbf{64.1} & \textbf{64.2} & \textbf{61.5} & 61.2 & 60.4 & \textbf{59.9} & 57.0 & 56.7 & 55.9 & 50.4\\
		SeqPtr\_CSG(Enc) & 63.5 & 62.9 & 62.9 & \textbf{61.5} & 62.1 & \textbf{61.1} & 56.8 & 58.0 & \textbf{57.0} & 55.4 & 51.2\\
		SeqPtr\_CSG(Sum) & \textbf{64.8} & 63.4 & 63.0 & 60.9 & 61.8 & 60.0 & 58.9 & 58.3 & 56.9 & \textbf{56.4} & \textbf{52.3}\\
		SeqPtr\_CSG(Cat) & 64.0 & 63.6 & 63.8 & 60.8 & \textbf{62.2} & 60.7 & 58.8 & \textbf{59.6} & 56.5 & 54.9 & 48.3\\
		\hline
		HD           & \textbf{66.3} & \textbf{66.7} & 64.6 & \textbf{65.0} & 61.6 & 61.2 & 57.3 & 59.5 & 57.4 & 55.1 & 48.4\\
		HD\_CSG(Enc) & 65.9 & 64.8 & 65.1 & 62.9 & 61.3 & 59.1 & 58.2 & 57.3 & 56.0 & 54.2 & 47.5\\
		HD\_CSG(Sum) & 65.6 & \textbf{66.7} & \textbf{65.8} & 64.2 & \textbf{63.2} & \textbf{61.9} & 57.1 & \textbf{59.6} & 57.8 & 54.6 & 48.9\\
		HD\_CSG(Cat) & 65.8 & 65.6 & 64.4 & 63.0 & 62.8 & 59.6 & \textbf{59.4} & 58.5 & \textbf{58.0} & \textbf{56.5} & \textbf{49.7}\\
		\hline
		TRADE           & 65.8 & 66.0 & \textbf{66.6} & \textbf{65.8} & 62.6 & 60.4 & \textbf{62.8} & 59.7 & 59.1 & \textbf{58.1} & 51.0\\
		TRADE\_CSG(Enc) & 65.7 & 64.9 & 65.4 & 64.1 & 62.4 & 58.3 & 60.2 & 59.1 & 56.9 & 56.9 & 50.0\\
		TRADE\_CSG(Sum) & \textbf{67.4} & \textbf{67.4} & \textbf{66.6} & 65.1 & 63.7 & \textbf{62.2} & 61.3 & \textbf{61.9} & \textbf{59.6} & 57.3 & \textbf{52.7}\\
		TRADE\_CSG(Cat) & 65.3 & 66.1 & 65.8 & 65.1 & 63.7 & 60.0 & 61.5 & 59.8 & \textbf{59.6} & 57.1 & 52.1\\
		\hline
	\end{tabular}
\end{table*}

\begin{figure*}[h]
	\centering
	\includegraphics[width=0.7\linewidth]{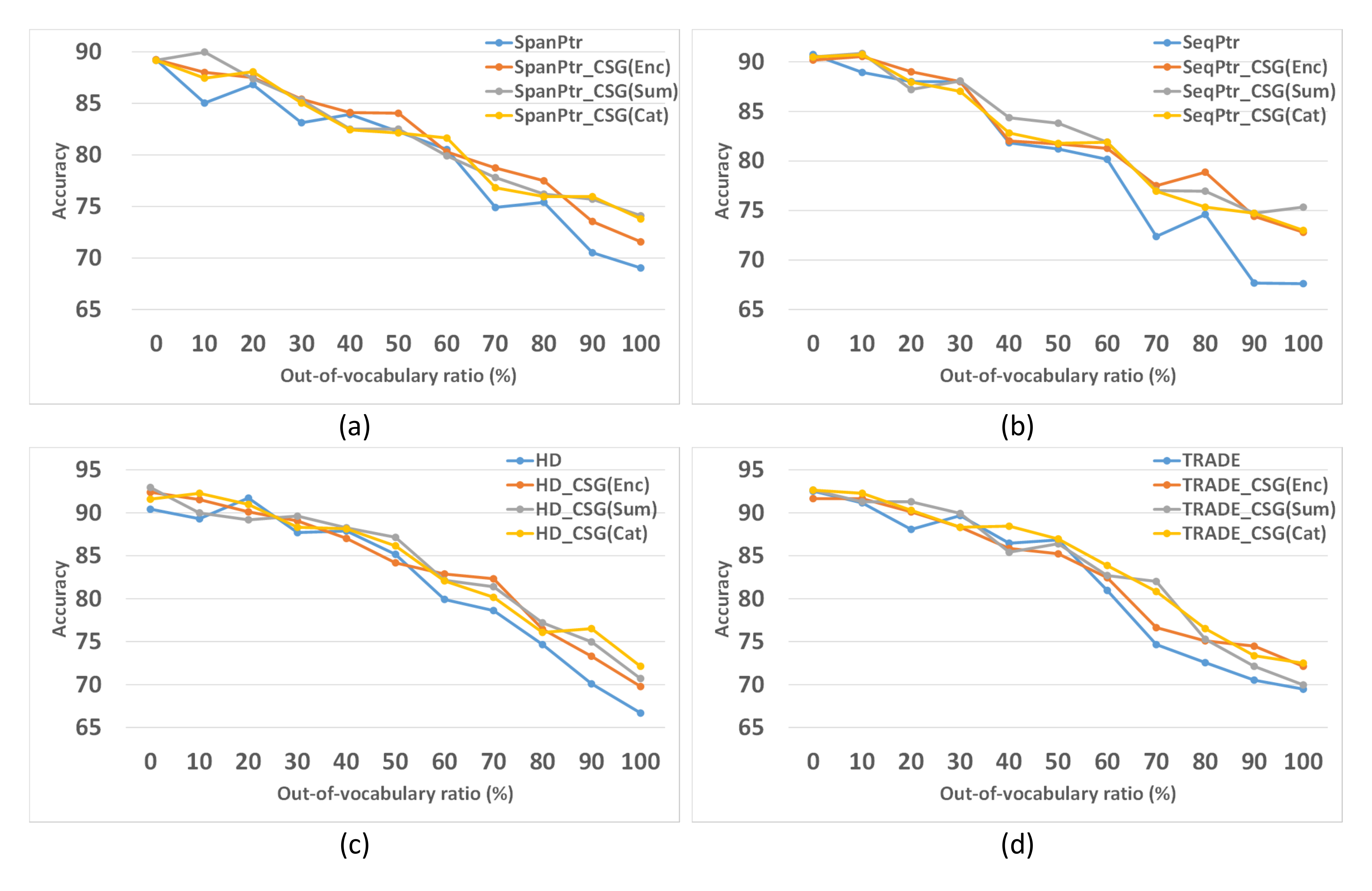}
	\caption{Joint accuracy of dialog state tracking in domain restaurant with different out-of-vocabulary ratios on
		modified MultiWOZ 2.1 dataset.}
	%\Description{The 1907 Franklin Model D roadster.}
	\label{img3}
\end{figure*}

To verify the performance of our model on the existing public datasets (there is basically no unknown slot value), we conduct experiments on the original MultiWOZ 2.1, DSTC2 datasets and the modified MultiWOZ 2.1 dataset. The joint accuracy of dialogue state tracking is shown in Table \ref{table3}. We can observe that, on the whole, our models are comparable to and sometimes even better than these baselines, especially on the DSTC2 dataset. Among them, the most significant improvement come from our HD\_CSG(Cat) model on the DSTC2 dataset, with an absolute improvement of 3.63.

By observing the performance of our model at each baseline in Table \ref{table3} we can find that the improvement effect of our model is different for different baselines. Specifically, SpanPtr\_CSG(Enc), SeqPtr\_CSG(Cat), HD\_CSG(Cat) and TRADE\_CSG(Sum) in our models are relatively optimal improvements at each baseline. Similarly, our model has different advantages in different datasets. In general, it can be seen that for different baselines on different datasets, we can always achieve results that outperform the baseline after utilizing the capabilities of our model.

In addition to ensuring the performance of the model on the existing public datasets, the ability to extract unknown slot values is the main focus of this paper. The joint accuracy of dialog state tracking (DST) on the modified MultiWOZ 2.1 dataset in different out-of-vocabulary ratios is shown in Table \ref{table4}. It should be emphasized here that the proposed model is mainly for the handing of unknown slot value containing multiple out-of-vocabulary words. In addition, since the MultiWOZ 2.1 dataset is target at complex DST in multiple domains, the final joint accuracy is not only relate to the extraction of unknown slot values, but also depend to other factors, such as cross-domain learning. Under this premise, according to Table \ref{table4}, in general, our model performs as well as all baselines when there are less than 38\% unknown slot values containing multiple out-of-vocabulary words (out-of-vocabulary ratio is less than 70\%). And when there are more than 38\% unknown slot value with multiple out-of-vocabulary words (the out-of-vocabulary ratio is greater than 70\%), our model is almost always outperforming all baselines. What can also be observed is that compared with pointer network-based extractive DST, the context utilization scheme ”Enc” does not perform very well on pointer-generator networks-based hybrid DST, this should be related to the fact that in complex multi-domain DST tasks, using only context information as input reduces the amount of information pointer-generator networks receives when generating vocabulary-based distributions.

The consistency of the ability to extract known slot values and unknown slot values is also the key to evaluate the practicability of the DST model. From the comprehensive assessment of the most prominent models in Table \ref{table3} and Table \ref{table4}, we can see that, our model basically maintains the consistency of the results on the original dataset and the dataset with different out-of-vocabulary ratios. For example, SpanPtr\_CSG(Enc) and TRADE\_CSG(Sum) have significant advantages over baseline in different datasets, greatly improving the modeling of unknown slot values while maintaining the ability to extract known slot values.

\begin{table*}[!t]
	%% increase table row spacing, adjust to taste
	%\renewcommand{\arraystretch}{1.3}
	% if using array.sty, it might be a good idea to tweak the value of
	% \extrarowheight as needed to properly center the text within the cells
	\caption{The accuracy in extracting slot values of different types on the modified MultiWOZ 2.1 dataset of individual restaurant domain with an out-of-vocabulary ratio of 70\%. KSV is the knowable slot value of all words in the word vacabulary, USV-O refers to the unknown slot value containing only one out-of-vocabulary word, while USV-M refers to the unknown slot
		value containing two or more out-of-vocabulary words.}
	\label{table5}
	\centering
	%% Some packages, such as MDW tools, offer better commands for making tables
	%% than the plain LaTeX2e tabular which is used here.
	\begin{tabular}{|c|c|c|c|}
		\hline
		& KSV & USV-O & USV-M \\
		\hline
		SpanPtr           & 90.36 & 77.30 & 66.36 \\
		SpanPtr\_CSG(Enc) & 89.16 (-1.20) & 81.89 (+4.59) & 71.27 (+4.91) \\
		SpanPtr\_CSG(Sum) & 87.95 (-2.41) & 79.67 (+2.37) & 71.89 (+5.53) \\
		SpanPt\_CSG(Cat)  & 89.56 (-0.80) & 80.78 (+3.48) & 67.59 (+1.23) \\
		\hline
		SeqPtr           & 89.56 & 78.41 & 59.14 \\
		SeqPtr\_CSG(Enc) & 89.16 (-0.40) & 81.89 (+3.48) & 68.20 (+9.06) \\
		SeqPtr\_CSG(Sum) & 89.16 (-0.40) & 79.53 (+1.12) & 69.59 (+10.45) \\
		SeqPtr\_CSG(Cat) & 84.34 (-5.22) & 81.89 (+3.48) & 68.66 (+9.52) \\
		\hline
		HD           & 84.58 & 86.36 & 67.30 \\
		HD\_CSG(Enc) & 83.70(-0.88) & 86.75(+0.39) & 76.57(+9.27) \\
		HD\_CSG(Sum) & 85.46(+0.88) & 87.02(+0.66) & 73.27(+5.97) \\
		HD\_CSG(Cat) & 86.78(+2.20) & 85.17(-1.19) & 71.86(+4.56) \\
		\hline
		TRADE           & 85.54 & 80.64 & 63.90 \\
		TRADE\_CSG(Enc) & 89.56 (+4.02) & 83.84 (+3.20) & 63.75 (-0.15) \\
		TRADE\_CSG(Sum) & 91.24 (+5.70) & 87.71 (+7.07) & 72.01 (+8.11) \\
		TRADE\_CSG(Cat) & 89.40 (+3.86) & 81.83 (+1.19) & 76.73 (+12.83) \\
		\hline
	\end{tabular}
\end{table*}

\begin{figure*}[h]
	\centering
	\includegraphics[width=0.7\linewidth]{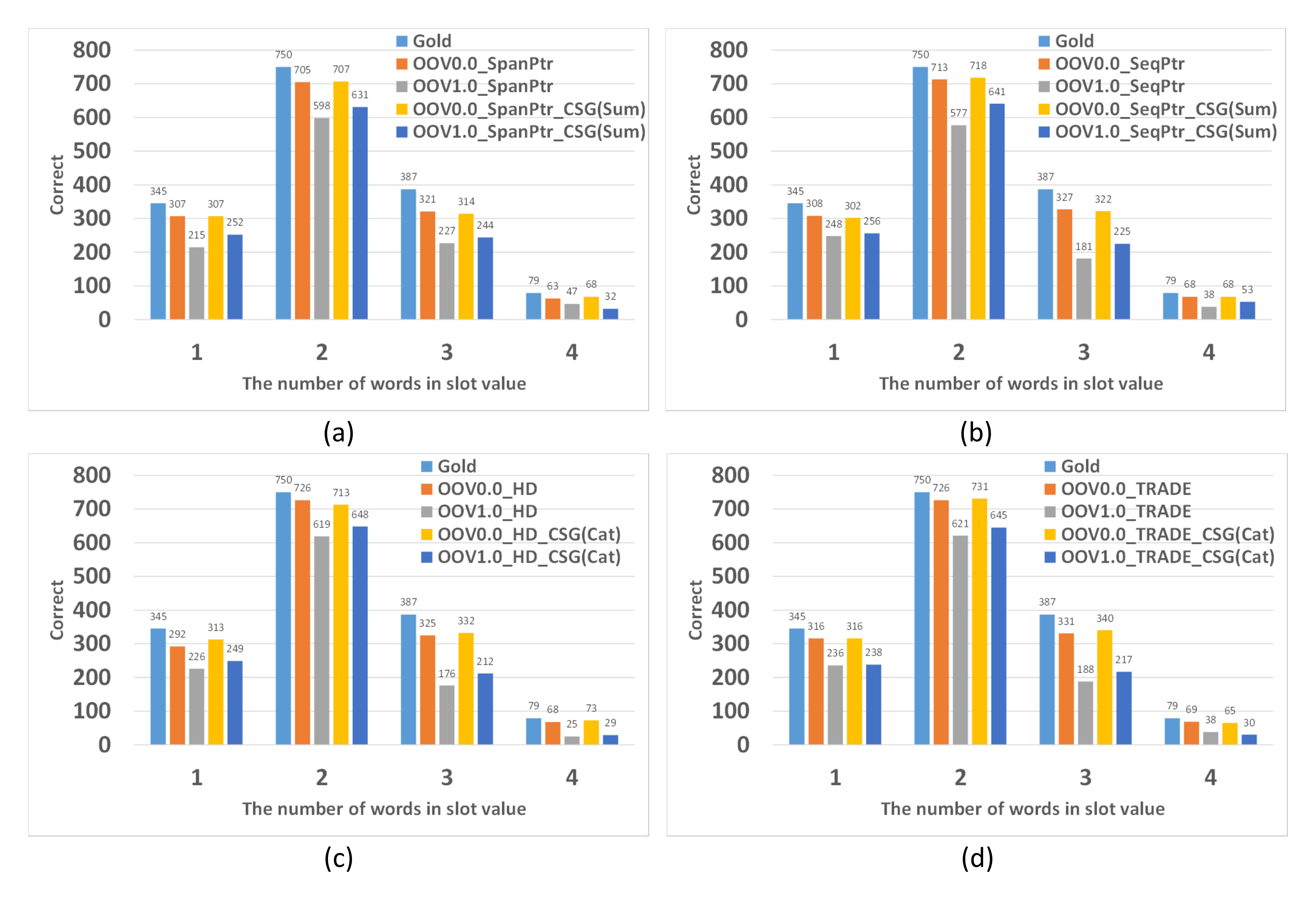}
	\caption{Comparison of the model’s correct predictions when slot values contains different numbers of words
		in domain restaurant on modified MultiWOZ 2.1 dataset. Gold is the ground truth result in the dataset, OOV0.0 refers to the out-of-vocabulary ratio of 0\%, where the unknown slot value is not included. OOV1.0 refers to an out-of-vocabulary ratio of 100\%, in which case all slot values are unknown slot values.}
	%\Description{The 1907 Franklin Model D roadster.}
	\label{img4}
\end{figure*}

The experiments in the individual restaurant domain can better highlight the superiority of our model over all baselines, as shown in Fig \ref{img3}, where the performance of all DST models is unaffected by knowledge sharing across domains. Here, we can observe more clearly that when out-of-vocabulary ratio exceeds 70\% (the proportion of unknown slot values containing multiple out-of-vocabulary words exceeds 38\%), all baselines are defective in extracting unknown slot values compared with our models. Besides, for pointer network-based extractive DST, the advantages of our model over the baseline are obviously gradually expanded with the proportion of out-of-vocabulary increasing, especially in the SeqPtr series models. This phenomenon shows that the context information plays a more important role in the pointer network-based extractive DST, which is inseparable from the fact that the pointer network-based extractive DST mainly relies on the dialogue context to extract the slot value.

\subsection{Analysis}

\begin{table}[!t]
	%% increase table row spacing, adjust to taste
	%\renewcommand{\arraystretch}{1.3}
	% if using array.sty, it might be a good idea to tweak the value of
	% \extrarowheight as needed to properly center the text within the cells
	\caption{The joint accuracy, slot accuracy and slot F1 of each model on the test set When 1\% of the training set is used in individual restaurant domain on the MultiWOZ 2.1 dataset.}
	\label{table6}
	\centering
	%% Some packages, such as MDW tools, offer better commands for making tables
	%% than the plain LaTeX2e tabular which is used here.
	\begin{tabular}{|c|c|c|c|}
		\hline
		& Joint Acc & Slot Acc & Slot F1 \\
		\hline
		SpanPtr           & 17.89 & 62.62 & 47.74 \\
		SpanPtr\_CSG(Enc) & 18.58 & \textbf{69.62} & \textbf{56.85} \\
		SpanPtr\_CSG(Sum) & \textbf{19.11} & 65.30 & 52.16 \\
		SpanPt\_CSG(Cat)  & 17.71 & 68.12 & 54.21 \\
		\hline
		SeqPtr           & 15.78 & 50.49 & 15.91 \\
		SeqPtr\_CSG(Enc) & \textbf{19.05} & \textbf{66.13} & \textbf{51.54} \\
		SeqPtr\_CSG(Sum) & 15.81 & 50.50 & 15.87 \\
		SeqPtr\_CSG(Cat) & 15.81 & 50.47 & 15.81 \\
		\hline
		HD           & \textbf{15.81} & 50.43 & 16.02 \\
		HD\_CSG(Enc) & \textbf{15.81} & 50.42 & 15.84 \\
		HD\_CSG(Sum) & \textbf{15.81} & 50.50 & 15.97 \\
		HD\_CSG(Cat) & 15.78 & \textbf{51.84} & \textbf{21.48} \\
		\hline
		TRADE           & 15.75 & 50.61 & 16.97 \\
		TRADE\_CSG(Enc) & 15.81 & 50.47 & 15.81 \\
		TRADE\_CSG(Sum) & 15.81 & 50.47 & 15.81 \\
		TRADE\_CSG(Cat) & \textbf{16.02} & \textbf{53.41} & \textbf{26.45} \\
		\hline
	\end{tabular}
\end{table}

In individual restaurant domain data with an out-of-vocabulary ratio of 70\% (KSV 14\%, USV-O 47\%, USV-M 39\%), the accuracy of each model in extracting slot values of different types is shown in Table 5. (1) For KSV, our model is basically perform as well as the baseline model. On the one hand, for the extractive
DST, our model is slightly lower than the baseline model on the whole, with the largest reduction coming from SeqPtr\_CSG(Cat). On the other hand, for the hybrid DST, our model basically achieve an overall improvement, with the maximum increase reaching 5.70, which indicates that context information is not only beneficial to USV extraction, but also can promote the generation of KSV in pointer-generator networks to some extent. (2) For USV-O, although the baseline model can retain some extraction capability with the help of pointer mechanism, the results show a significant decrease compared with KSV except for the HD series model. In contrast, our model generally achieves exciting results, such as TRADE\_CSG(Sum) with a 7.07 improvement, which reflects our model's better modeling of all types of USVs, not just USV-M. (3) For USV-M, our model achieve remarkable results, while all baseline models have a significant, unavoidable drop, which is close to 26\% on average compared to KSV extraction. Meanwhile, All of our models better guarantee the extraction of USV-M, and the maximum absolute improvement reach to 12.83 compared with the baseline model. In general, with the help of word contextual representation, our model enriches the out-of-vocabulary word representation and avoids the problem of insufficient out-of-vocabulary word information caused by only using vocabulary word embedding. Therefore, it effectively solves the problem of the traditional model's insufficient ability to extract USV, especially the extraction of USV-M.

The influence of the number of out-of-vocabulary words on the extraction of unknown slot values is shown in Fig \ref{img4}. The “Sum” scheme on the extractive DST model and the “Cat” scheme on the hybrid DST model were compared with the baseline. It can be observed that as the number of out-of-vocabulary words in unknown slot value increases, the difficulty of unknown slot value extraction also increases gradually, and when there are more than three out-of-vocabulary words in the unknown slot value, the accuracy of the baseline model to extract the unknown slot value is almost less than 50\%, while in contrast, the accuracy of our model is close to 60\%, which is a significant improvement. In addition, a more exciting phenomenon we can find is that as the number of out-of-vocabulary words in the unknown slot value increases, the advantages of our model over the baseline model gradually expands, which can be seen from the improvement of the relative accuracy of our model to the baseline model under different number of out-of-vocabulary words. In general, our model improves the ability to extract the USV containing multiple OOV words, especially for those USVs containing more out-of-vocabulary words in which the traditional pointer mechanism-based model is almost invalid.

As we all know, in the case of insufficient training data, the model often faces a large number of USVs, which is a big obstacle to the generalization ability of the model. Therefore, we evaluate the performance of the model in the extremely lack of training data, as shown in Table \ref{table6}. We can clearly find that when the model is faced with an extreme lack of training data, our model can relatively improve the generalization ability of the model. For the two types of DST models, the impact of our model is different: (1) For extractive DST, our proposed "Enc" scheme (word contextual representation only) achieves the best results, which shows that when there is not enough training data to construct an adequate vocabulary space, the use of context to represent each word is more in line with the modeling requirements of the model for dialogue history distribution. (2) For hybrid DST, since the model needs to solve the vocabulary-based distribution and  dialogue history distribution respectively and combine them when generating slot values, the "Cat" scheme achieves better results. This is mainly due to the fact that the "Cat" scheme combines the word vocabulary embedding and context representation, which provides more abundant information for solving the two distributions.

\section{Conclusion}

In this paper, we point out the defects of the current pointer mechanism-based dialogue state tracking model in extracting unknown slot values, and propose a novel model to extract unknown slot values more effectively by enhancing the representation of word with the word contextual information, namely, context-sensitive generation network (CSG). We also propose three different context utilization schemes for the CSG: (1) "Enc", The contextual representation directly replaces the word embedding, (2) "Sum", The contextual representation is summed with the word embedding, (3) "Cat", the concatenation of the word embedding and contextual representations. Extensive experiments on the MultiWOZ 2.1, DSTC2 and our modified MultiWOZ 2.1 dataset show that, compared with the existing baseline models, our proposed model not only retains the extraction of knowable slot values, but also greatly improves the processing capacity of unknown slot values, especially for the unknown slot value containing multiple out-of-vocabulary words. Meanwhile, better modeling of unknown slot value also enables our proposed model to have low resource learning capability.

% use section* for acknowledgment
\ifCLASSOPTIONcompsoc
  % The Computer Society usually uses the plural form
  \section*{Acknowledgments}
\else
  % regular IEEE prefers the singular form
  \section*{Acknowledgment}
\fi

The authors would like to thank...

% Can use something like this to put references on a page
% by themselves when using endfloat and the captionsoff option.
\ifCLASSOPTIONcaptionsoff
  \newpage
\fi

% trigger a \newpage just before the given reference
% number - used to balance the columns on the last page
% adjust value as needed - may need to be readjusted if
% the document is modified later
%\IEEEtriggeratref{8}
% The "triggered" command can be changed if desired:
%\IEEEtriggercmd{\enlargethispage{-5in}}

% references section

% can use a bibliography generated by BibTeX as a .bbl file
% BibTeX documentation can be easily obtained at:
% http://mirror.ctan.org/biblio/bibtex/contrib/doc/
% The IEEEtran BibTeX style support page is at:
% http://www.michaelshell.org/tex/ieeetran/bibtex/
%\bibliographystyle{IEEEtran}
% argument is your BibTeX string definitions and bibliography database(s)
%\bibliography{IEEEabrv,../bib/paper}
%
% <OR> manually copy in the resultant .bbl file
% set second argument of \begin to the number of references
% (used to reserve space for the reference number labels box)
%\begin{thebibliography}{1}

%\bibitem{IEEEhowto:kopka}
%H.~Kopka and P.~W. Daly, \emph{A Guide to \LaTeX}, 3rd~ed.\hskip 1em plus
%  0.5em minus 0.4em\relax Harlow, England: Addison-Wesley, 1999.
\bibliographystyle{IEEEtran}
\bibliography{TKDE}

%\end{thebibliography}

% biography section
% 
% If you have an EPS/PDF photo (graphicx package needed) extra braces are
% needed around the contents of the optional argument to biography to prevent
% the LaTeX parser from getting confused when it sees the complicated
% \includegraphics command within an optional argument. (You could create
% your own custom macro containing the \includegraphics command to make things
% simpler here.)
%\begin{IEEEbiography}[{\includegraphics[width=1in,height=1.25in,clip,keepaspectratio]{mshell}}]{Michael Shell}
% or if you just want to reserve a space for a photo:

\iffalse

\begin{IEEEbiography}{Puhai Yang}
Biography text here.
\end{IEEEbiography}

% if you will not have a photo at all:
\begin{IEEEbiographynophoto}{Heyan Huang}
Biography text here.
\end{IEEEbiographynophoto}

% insert where needed to balance the two columns on the last page with
% biographies
%\newpage

\begin{IEEEbiographynophoto}{Xian-Ling Mao}
Biography text here.
\end{IEEEbiographynophoto}
\fi

% You can push biographies down or up by placing
% a \vfill before or after them. The appropriate
% use of \vfill depends on what kind of text is
% on the last page and whether or not the columns
% are being equalized.

%\vfill

% Can be used to pull up biographies so that the bottom of the last one
% is flush with the other column.
%\enlargethispage{-5in}

% that's all folks
\end{document}